\documentclass[10pt,twocolumn,letterpaper]{article}

\usepackage{iccv}              

\usepackage[belowskip=-7pt,aboveskip=3pt]{caption}

\usepackage{multirow}
\usepackage{multibib}
\newcites{supp}{Supplementary References}
\newif\ifdrafting
\draftingtrue 
\ifdrafting

\else

\fi

\definecolor{guppiegreen}{rgb}{0.0, 0.7, 0.7}
\definecolor{alizarin}{rgb}{0.8, 0.1, 0.2}
\definecolor{brandeisblue}{rgb}{0.0, 0.44, 1.0}

\newcommand{\st}[1]{\textbf{{\color{red}#1}}}
\newcommand{\bd}[1]{\textbf{{\color{blue}#1}}}

\definecolor{iccvblue}{rgb}{0.21,0.49,0.74}
\usepackage[pagebackref,breaklinks,colorlinks,allcolors=iccvblue]{hyperref}

\def\paperID{2221} 
\def\confName{ICCV}
\def\confYear{2025}

\title{EAMamba: Efficient All-Around Vision State Space Model \\ for Image Restoration}

\author{
Yu-Cheng Lin\textsuperscript{*, 1, 3} \ Yu-Syuan Xu\textsuperscript{*, 2, 3} \ Hao-Wei Chen\textsuperscript{2} \ Hsien-Kai Kuo\textsuperscript{3} \ Chun-Yi Lee\textsuperscript{1, 2}
\smallskip
\\
\textsuperscript{1}National Tsing Hua University \quad 
\textsuperscript{2}National Taiwan University \quad 
\textsuperscript{3}MediaTek Inc. \\
{\tt\small \{lyc1112, jaroslaw1007\}@gapp.nthu.edu.tw \quad}
{\tt\small cylee@csie.ntu.edu.tw} \\
{\tt\small \{Yu-Syuan.Xu, Hsienkai.Kuo\}@mediatek.com} \\
}

\newcommand\nnfootnote[1]{%
  \begin{NoHyper}
  \renewcommand\thefootnote{}\footnote{#1}%
  \addtocounter{footnote}{-1}%
  \end{NoHyper}
}

\begin{document}
\maketitle
\nnfootnote{*Equal contribution}

\vspace{-20pt}
\begin{abstract}
\vspace{-20pt}

Image restoration is a key task in low-level computer vision that aims to reconstruct high-quality images from degraded inputs. The emergence of Vision Mamba, which draws inspiration from the advanced state space model Mamba, marks a significant advancement in this field. Vision Mamba demonstrates excellence in modeling long-range dependencies with linear complexity, a crucial advantage for image restoration tasks. Despite its strengths, Vision Mamba encounters challenges in low-level vision tasks, including computational complexity that scales with the number of scanning sequences and local pixel forgetting. To address these limitations, this study introduces Efficient All-Around Mamba (EAMamba), an enhanced framework that incorporates a Multi-Head Selective Scan Module (MHSSM) with an all-around scanning mechanism. MHSSM efficiently aggregates multiple scanning sequences, which avoids increases in computational complexity and parameter count. The all-around scanning strategy implements multiple patterns to capture holistic information and resolves the local pixel forgetting issue. Our experimental evaluations validate these innovations across several restoration tasks, including super resolution, denoising, deblurring, and dehazing. The results validate that EAMamba achieves a significant 31-89\% reduction in FLOPs while maintaining favorable performance compared to existing low-level Vision Mamba methods. The source codes are accessible at the following repository: \href{https://github.com/daidaijr/EAMamba}{https://github.com/daidaijr/EAMamba}. 

\vspace{-20pt}
\end{abstract}

\section{Introduction}
\label{sec:intro}

\DeclareRobustCommand{\ourmethod}{\textcolor[RGB]{240, 5, 23}{\raisebox{-0.1ex}{\scalebox{1.2}{$\bigstar$}}}}
\DeclareRobustCommand{\othermethod}{\textcolor[RGB]{37, 42, 223}{\raisebox{-0.4ex}{\scalebox{2.0}{$\bullet$}}}}
\DeclareRobustCommand{\existingmethod}{\textcolor[RGB]{41, 165, 44}{\raisebox{-0.4ex}{\scalebox{2.0}{$\bullet$}}}}

\begin{figure}[t]
    \centering
    \footnotesize
    \vspace{-10pt}
    \begin{minipage}[t]{0.45\columnwidth}
        \centering
        \includegraphics[width=\columnwidth]{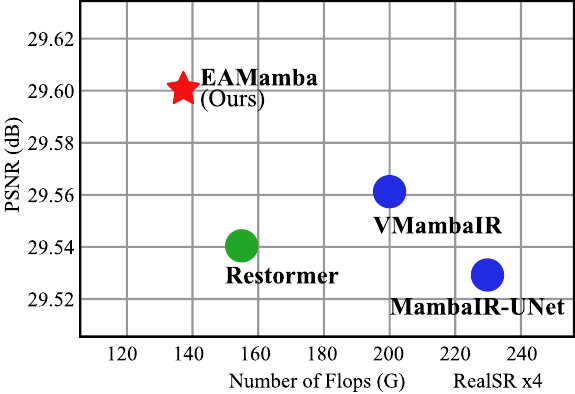}
        \centerline{\qquad \footnotesize (a) Real-World Super-Resolution}
    \end{minipage}
    \begin{minipage}[t]{0.45\columnwidth}
        \centering
        \includegraphics[width=\columnwidth]{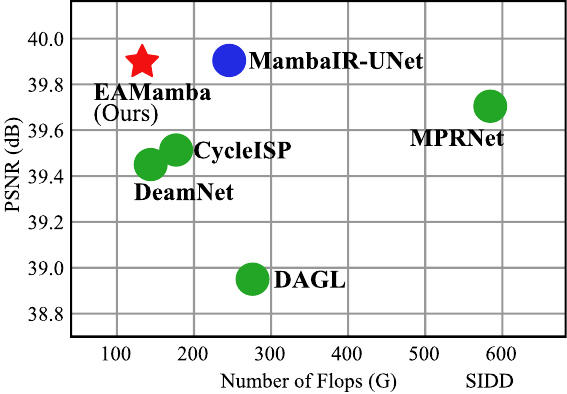}
        \centerline{\qquad \footnotesize (b) Real-World Denoising}
    \end{minipage}
    \\
    \begin{minipage}[t]{0.45\columnwidth}
        \centering
        \includegraphics[width=\columnwidth]{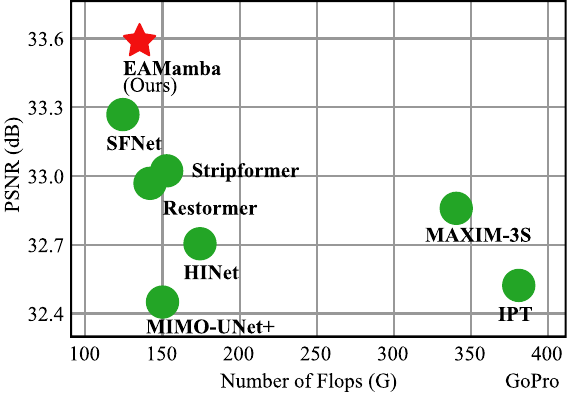}
        \centerline{\qquad \footnotesize (c) Motion Deblurring}
    \end{minipage}
    \begin{minipage}[t]{0.45\columnwidth}
        \centering
        \includegraphics[width=\columnwidth]{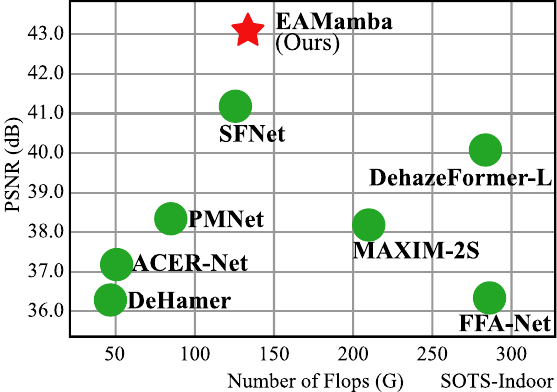}
        \centerline{\qquad \footnotesize (d) Indoor Dehazing}
    \end{minipage}
    \caption{
        Computational efficiency versus image quality across model architectures. Our method (denoted by \ourmethod) demonstrates superior efficiency compared to other Vision Mamba-based methods (\othermethod) and existing approaches (\existingmethod). EAMamba establishes a new efficiency frontier for Vision Mamba-based image restoration.
    }
    \label{fig:compare}
    \vspace{-15pt}
\end{figure}

Image restoration constitutes a crucial task in computer vision and image processing. The primary objective of this task is to reconstruct high-quality images from low-quality inputs affected by various degradations such as noise, blurring, and downsampling. Due to the underlying ill-posed nature of image restoration, this task presents substantial technical challenges. In early developments, Convolutional Neural Network (CNN)-based approaches have demonstrated success across various image restoration benchmarks~\cite{srcnn, vdsr, edsr, srresnet, esrgan, rcan, rdn, cbdnet, dncnn, ircnn, ffdnet, brdnet, drunet, deblurgan, srn, deblurganv2, dbgan, dmphn, mimo, derainnet, semi, didmdn, umrl, rescan, prenet, mspfn, mscnn, dehazenet, aodnet, griddehazenet, msbdn, ffanet, acernet, pmnet, mirnet, mprnet, hinet, spair, sfnet}. However, these architectures exhibit inherent limitations in their ability to capture global information due to their focus on local pixel relationships. Subsequent research explored Vision Transformer (ViT)~\cite{vit}-based architectures, which employ multi-head self-attention mechanisms to model relationships across all image pixels. 
These approaches effectively enable the capture of global dependencies and have achieved promising results across multiple restoration tasks~\cite{san, hat, stripformer, idt, drsformer, dehamer, dehazeformer, swinir, ipt, uformer, restormer, art}. Nevertheless, the computational complexity of the attention mechanisms used scales quadratically with pixel count, rendering high-resolution image processing infeasible and motivating further investigation into more efficient architectural designs.

\begin{figure}[t]
    \centering
    \includegraphics[width=0.9\columnwidth]{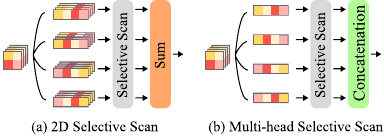}
    \caption{Illustration of architectural differences: two-dimensional Selective Scan (2DSS) operating on full feature channels versus our Multi-head Selective Scan (MHSS) utilizing feature splitting.}
    \label{fig:efficient}
    \vspace{-10pt}
\end{figure}

\begin{figure}[t]
    \centering
    \includegraphics[width=0.9\columnwidth]{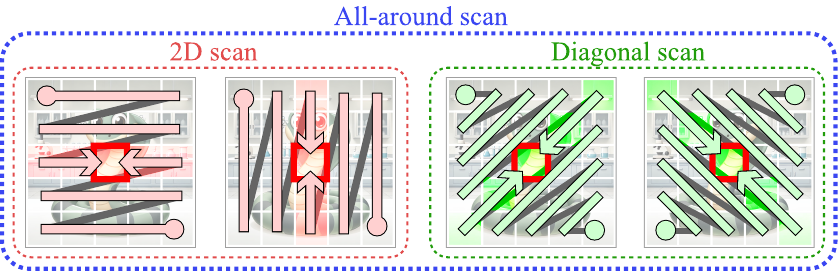}
    \caption{Illustration of an all-around scanning approach that combines two-dimensional scanning and diagonal scanning.}
    \label{fig:all-around}
    \vspace{-5pt}
\end{figure}

In light of the computational complexity challenges, the recently introduced Mamba framework~\cite{mamba}, an advanced state space model, presents a promising solution. This framework demonstrates linear computational scaling characteristics and efficient long-range dependency modeling capabilities, and has shown exceptional potential in natural language processing (NLP) tasks. The potential has led to its adaptation for vision-related tasks through Vision Mamba models~\cite{vmamba, vim, mambavision, mambair, vmambair} that aim to efficiently capture global image information while maintaining linear computational complexity with respect to pixel count. Vision Mamba adopts patch processing techniques similar to ViT and processes these patches through a state space model with a two-dimensional selective scanning strategy. Specifically, previous seminal works~\cite{vmamba, vim} have proposed the transformation of two-dimensional feature maps into flattened one-dimensional sequences through a dual-direction scanning process that incorporates both horizontal and vertical scans, as illustrated in Fig.~\ref{fig:all-around}. However, such methods introduce drawbacks: both computational complexity and parameter count scale linearly with the number of flattened one-dimensional sequences. As depicted in Fig.~\ref{fig:efficient}~(a), two-dimensional Selective Scan (2DSS) generates four one-dimensional sequences through different scanning patterns. Each sequence requires selective scanning with distinct parameters before combination into the final output, which inevitably increases computational overhead when incorporating additional scanning directions. As a result, the realization of enhanced scanning patterns would necessitate an alternative design that can ensure computational efficiency.

Motivated by the aforementioned challenges, this study introduces Efficient All-Around Mamba (EAMamba), an innovative Vision Mamba framework that integrates holistic scanning capabilities through a Multi-Head Selective Scan Module (MHSSM). The proposed Multi-Head Selective Scan (MHSS), as illustrated in Fig.~\ref{fig:efficient}~(b), employs channel grouping strategies for selective scanning operations, with subsequent concatenation of group outputs to produce comprehensive feature representations. A fundamental advantage of MHSS lies in its efficient processing of multiple one-dimensional sequences through the elimination of computational and parameter overhead typically associated with increased sequence numbers. This enhancement improves both the scalability of the Vision Mamba framework across various scan configurations as well as the capacity for efficient processing of more complex and diverse visual data.

\begin{figure}[t]
    \centering
    \includegraphics[width=0.9\columnwidth]{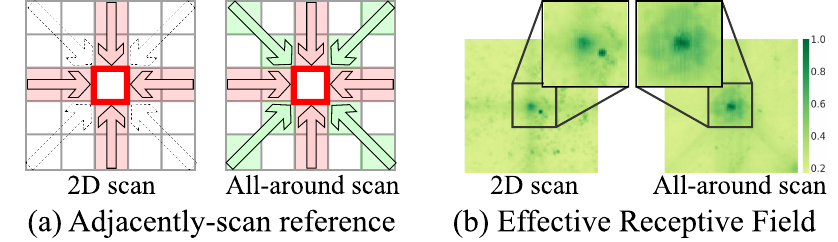}
    \caption{(a) Illustration of the local pixel forgetting phenomenon, where spatially adjacent pixels become distantly separated in the one-dimensional token sequence during scanning. The target pixel (highlighted in red square) and its adjacent pixels demonstrate how different scanning patterns affect spatial relationships. (b) The ERF visualization results averaged across the SIDD dataset~\cite{sidd}, which depict improved spatial dependency preservation with the proposed all-around scanning approach.}
    \label{fig:local-forgetting}
    \vspace{-10pt}
\end{figure}

In addition to computational considerations, the two-dimensional scanning approach presents another critical limitation: \textit{local pixel forgetting}, as mentioned in~\cite{mambair}. This phenomenon occurs when spatially adjacent pixels in the two-dimensional feature map become distantly separated in the one-dimensional token sequence, as illustrated in Fig.~\ref{fig:local-forgetting}~(a). Fig.~\ref{fig:local-forgetting}~(b) further depicts the extent of local information loss through an Effective Receptive Field (ERF) visualization that reveals each pixel's contribution to the target pixel output via gradient flow analysis. This analysis highlights how two-dimensional scanning can lead to local pixel forgetting. Such local information loss poses particular challenges for image restoration tasks, where preservation of local spatial relationships is crucial. Despite the importance of local spatial coherence, current scanning strategies have not adequately addressed this limitation, presenting significant opportunities for innovations in this domain.

To address the local pixel forgetting issue, EAMamba proposes a solution enhancement of scanning patterns for more comprehensive neighborhood information capture. This approach benefits from the efficiency of MHSS that enables EAMamba to incorporate holistic spatial information through a combination of multiple scans, termed \textit{all-around scanning} as shown in Fig.~\ref{fig:all-around}, to provide more complete neighborhood coverage. The motivation for adopting this strategy is supported by the ERF comparison illustrated in Fig.~\ref{fig:local-forgetting}~(b). It can be observed that two-dimensional scanning, as adopted in MambaIR~\cite{mambair}, faces difficulties in capturing information from diagonal directions, despite its supplementary local convolution operations. Our investigations suggest that incorporating additional scanning directions achieves a larger receptive field, particularly in preserving local information surrounding target pixels, which is a crucial for image restoration tasks.

\begin{figure*}[t]
    \centering
    \includegraphics[width=0.95\textwidth]{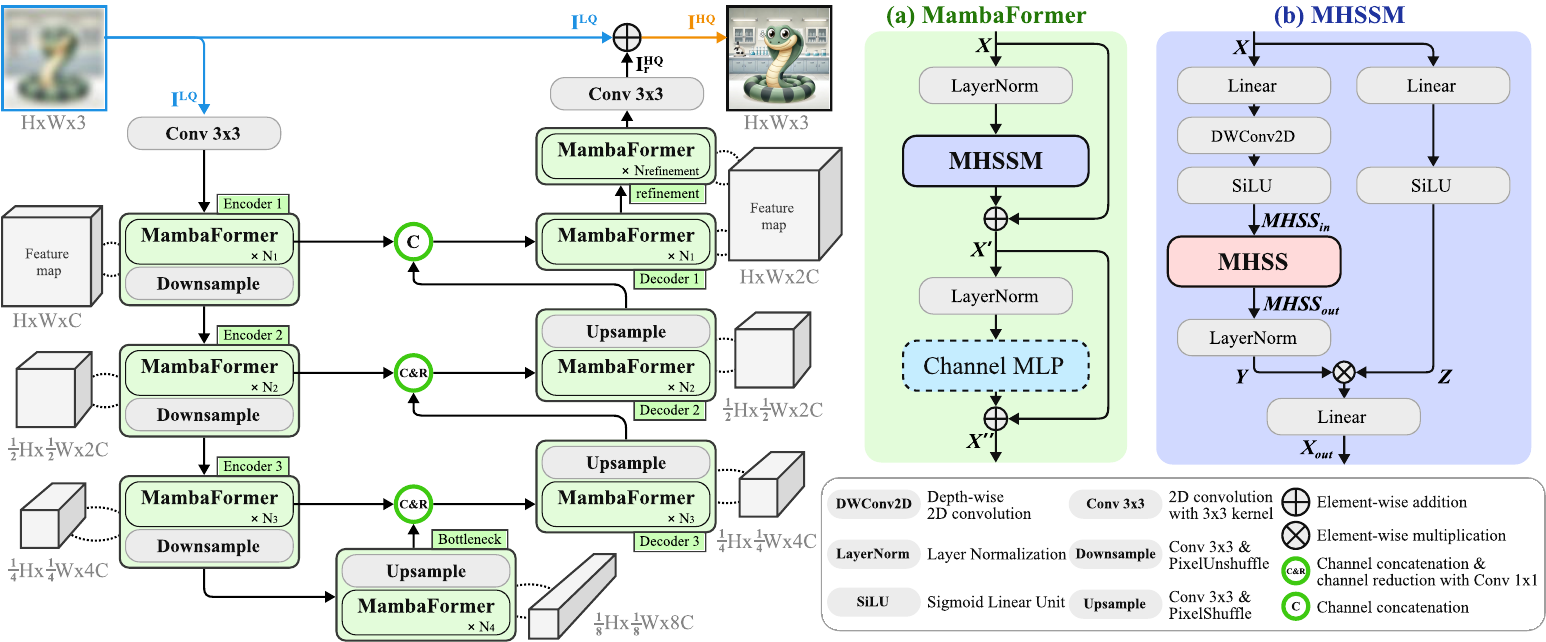}
    \caption{An overview of the proposed EAMamba framework. EAMamba framework is an encoder-decoder architecture. Both the encoder and the decoder are composed of multiple MambaFormer blocks organized across 4 levels, each captures features at different scales. (a) The MambaFormer block is inspired by the Transformer-based block~\cite{metaformer} that the token mixer is the proposed MHSSM. (b) The MHSSM consists of two branches: one for scanning, the other one for gating, which are ultimately merged through element-wise multiplication.}
    \label{fig:main-arch}
    \vspace{-10pt}
\end{figure*}

To validate EAMamba's effectiveness, we design extensive experiments across various types of image restoration tasks, including super-resolution (SR), denoising, deblurring, and dehazing. Our results demonstrate favorable performance on widely adopted benchmark datasets~\cite{realsr, sidd, gopro, reside} while most importantly requiring fewer parameters and lower FLOPs compared to existing low-level Vision Mamba methods. An illustrative comparison of computational efficiency and Peak Signal-to-Noise Ratio (PSNR) is presented in Fig.~\ref{fig:compare}. The key contributions are summarized as follows:
\begin{itemize}
    \item  We propose MHSSM to efficiently process and aggregate flattened 1D sequences without computational complexity and parameter count overhead, and improve the scalability and efficiency of the Vision Mamba frameworks.
    \item  We introduce an all-around scanning strategy that effectively captures holistic image information, addressing the issue of local pixel forgetting and enhancing the model's ability to understand spatial relationships within images.
    \item We carry out comprehensive analyses of the performance impacts for all-around scanning and MHSSM, and provide insights into the advantages of these innovations.
\end{itemize}

\section{Related Work}
\label{sec:bg}

\paragraph{Image Restoration.}
The field of image restoration covers multiple degradation types, each with its distinct challenges. Previous endeavors have targeted specific problems including super-resolution~\cite{srcnn, vdsr, edsr, srresnet, esrgan, rcan, rdn, san, hat, mma}, denoising~\cite{cbdnet, dncnn, ircnn, ffdnet, brdnet, drunet}, deblurring~\cite{deblurgan, srn, deblurganv2, dbgan, dmphn, mimo, stripformer}, deraining~\cite{derainnet, semi, didmdn, umrl, rescan, prenet, mspfn, idt, drsformer}, and dehazing~\cite{mscnn, dehazenet, aodnet, griddehazenet, msbdn, ffanet, acernet, pmnet, dehamer, dehazeformer}. This work investigates a unified Vision Mamba-based restoration framework that enhances reconstruction quality across multiple types of degradation.

\vspace{-12pt}
\paragraph{Vision Transformer.}
ViT~\cite{vit} has been a pivotal development in processing images, which utilizes self-attention mechanisms to grasp global dependencies. 
However, the application of ViTs in high-resolution image restoration is hampered by scalability issues due to their computational intensity. 
To overcome these limitations, researchers have introduced several innovative solutions. SwinIR~\cite{swinir} and Uformer~\cite{uformer} employ window-based attention mechanisms that enhance restoration quality while maintaining efficiency. 
Restormer~\cite{restormer} introduces multi-Dconv head transposed attention and gated-Dconv feed-forward to reduce computational complexity and improve performance.
Nonetheless, finding the balance between high performance and computational efficiency remains a critical challenge.

\vspace{-12pt}
\paragraph{Vision Mamba.}
Recent advancements in adapting Mamba techniques to vision tasks, as demonstrated by pioneering works~\cite{vim, vmamba}, have yielded promising outcomes.
The researchers in~\cite{vim} process the image patches in both forward and backward directions to better capture spatial information. In addition,
VMamba~\cite{vmamba} introduces cross-scan and cross-merge techniques to effectively aggregate spatial information.
Moreover, exploration into the application of Vision Mamba for image restoration tasks~\cite{mambair, vmambair} has revealed its considerable potential. 
MambaIR~\cite{mambair} integrates the vision state space module and the modified MLP block that mitigates local pixel forgetting and channel redundancy issues in vanilla Mamba. 
VMambaIR~\cite{vmambair} proposes an innovative module termed Omni Selective Scan (OSS), which conducts four-directional spatial scanning along with channel scanning to leverage both spatial and channel-wise information.
These advancements highlight Vision Mamba's capacity for capturing long-range dependencies and detailed contextual information within images, making it a promising solution for advanced vision tasks.

\section{Methodology}
\label{sec:method}
In this section, we first present an overview of the proposed EAMamba framework in Section~\ref{subsec::overview}. Subsequently, we provide a detailed examination of fundamental building blocks in Section~\ref{subsec::mambaformer}, and then elaborate on our proposed MHSSM and all-around scanning strategy in Section~\ref{subsec::mhssm}.

\subsection{Overview of the EAMamba Framework}
\label{subsec::overview}
To explore the potential of state space modeling and to address both computational overhead and the local pixel forgetting issue, we introduce EAMamba, a framework tailored for image restoration. Fig.~\ref{fig:main-arch} illustrates an overview of the proposed EAMamba framework that adopts a UNet-like architecture~\cite{unet}. The restoration process begins with a low-quality image $I^{LQ} \in \mathbb{R}^{H \times W \times 3}$. EAMamba processes this input through three MambaFormer encoder modules operating at different scales, followed by a bottleneck module. These stages extract feature embeddings with progressively varying dimensions, as illustrated in Fig.~\ref{fig:main-arch}. The framework then processes these features through three MambaFormer blocks for decoding and a refinement module to produce a residual image $I^{HQ}{r} \in \mathbb{R}^{H \times W \times 3}$. The final high-quality image $I^{HQ} \in \mathbb{R}^{H \times W \times 3}$ results from the element-wise addition of $I^{HQ}{r}$ and the original $I^{LQ}$. One of the primary contributions of EAMamba lies in the introduction of the MambaFormer architecture and MHSSM, which enable efficient scanning and incorporate holistic spatial information. These components are detailed in the following subsections. EAMamba demonstrates versatility over multiple image restoration tasks, which are presented in Section~\ref{sec:experiment}.

\subsection{MambaFormer Block}
\label{subsec::mambaformer}
MambaFormer blocks serve as integral components of EAMamba and are incorporated in the encoder, decoder, bottleneck, and refinement stages. As depicted in Fig.~\ref{fig:main-arch}~(a), each MambaFormer block comprises two primary components: the Multi-Head Selective Scan Module (MHSSM) for token mixing and a channel Multilayer Perceptron (channel MLP) for feature refinement. MambaFormer applies Layer Normalization (LN)~\cite{layernorm} before each component, with residual connections integrating the outputs into the preceding input:
\begin{equation} 
    \begin{aligned} 
        X' &= X + \text{MHSSM}(\text{LN}(X)), \\
        X'' &= X' + \text{Channel MLP}(\text{LN}(X')),
    \end{aligned}
\end{equation}
where $X$ denotes the input feature. In this architecture, MHSSM focuses on capturing long-range spatial dependencies within input features, while the channel MLP refines these features to enhance the representation. Section~\ref{subsec:ablation_studies} presents a comparison of different channel MLP designs.

\vspace{1pt}
\subsection{Multi-Head Selective Scan Module (MHSSM) with All-Around Scanning}
\vspace{1pt}
\label{subsec::mhssm}
MHSSM is a key component in MambaFormer that enhances the visual selective scan module introduced in~\cite{vmamba} through the replacement of two-dimensional Selective Scan (2DSS) with Multi-Head Selective Scan (MHSS), as detailed in Section~\ref{subsubsection::MHSS}. As illustrated in Fig.~\ref{fig:main-arch}~(b), the architecture processes input feature $X \in \mathbb{R}^{H \times W \times C}$ through two parallel branches. The left branch expands feature channels to $\lambda C$ through a linear layer, where $\lambda$ denotes a pre-defined channel expansion factor. The expanded features then pass through a sequence of operations: depth-wise convolution (DWConv2D), Sigmoid Linear Unit (SiLU)~\cite{silu}, MHSS, and LN. The right branch performs channel expansion by $\lambda$ followed by SiLU activation. The outputs from both branches combine through element-wise multiplication. A final linear projection reduces the merged output to the original dimension $C$. The complete MHSSM process is formulated as the following equations:
\begin{equation}
    \begin{aligned}
        Y &= \text{LN}( \text{MHSS}( \text{SiLU}(\text{DWConv2D}(\text{Linear}(X))) ) ), \\
        Z &= \text{SiLU}(\text{Linear}(X)),\quad X_{out} = \text{Linear}(Y \otimes Z),
    \end{aligned}
\end{equation}
where $X_{out} \in \mathbb{R}^{H \times W \times C}$ represents the output of the MHSSM block, and $\odot$ denotes element-wise multiplication.

\begin{figure}[t]
    \centering
    \includegraphics[width=\linewidth]{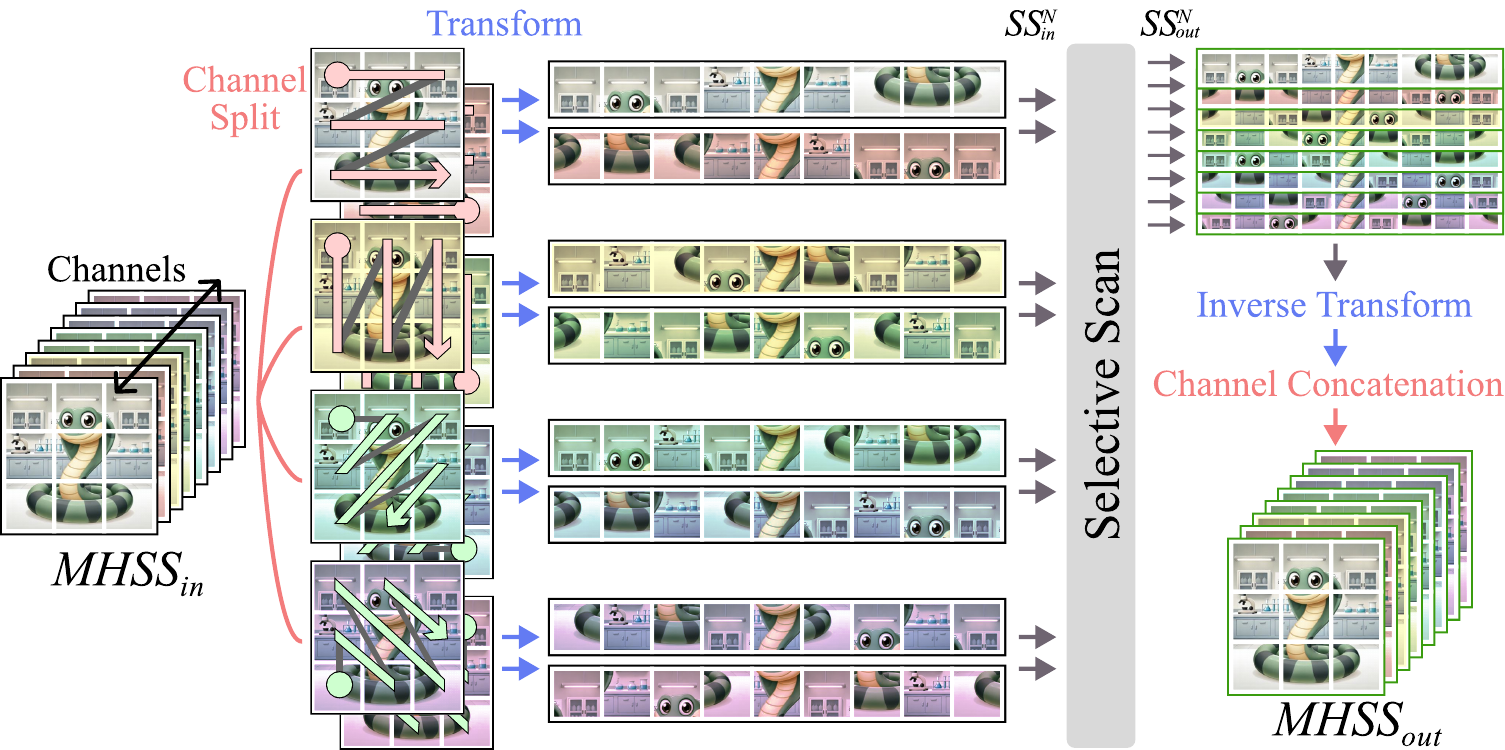}
    \caption{Illustration of the Multi-Head Selective Scan (MHSS) with our proposed All-Around Scanning strategy.}
    \label{fig:mhss}
\end{figure}

\vspace{3pt}
\subsubsection{Multi-Head Selective Scan (MHSS)}
\vspace{2pt}
\label{subsubsection::MHSS}
MHSS employs a multi-head approach for long-range spatial information capture through grouped feature processing and selective scanning. As illustrated in Fig.~\ref{fig:mhss}, MHSS partitions the input feature $MHSS_{in}$ into $n$ groups along the channel dimension, rather than scanning through the entire channel dimension. Each group receives a transformation for scanning two-dimensional input, followed by flattening into a one-dimensional sequence $SS_{in}^i$. In this work, we implement the proposed all-around scanning strategy, with transformation details outlined in Section~\ref{subsubsection::all-around}. The selective scanning process then generates corresponding outputs $SS_{out}^i$ for each sequence. After inverse transformation of each $SS_{out}^i$, the outputs concatenate to form the final output $MHSS_{out}$. The MHSS operations are formulated as:
\begin{equation}
    \begin{aligned}
        SS_{in}^N &= \text{Transform}( \text{Split}(MHSS_{in}) ), \\
        SS_{out}^N &= \text{SelectiveScan}(SS_{in}^N), \\
        MHSS_{out} &= \text{Concat}( \text{Inverse Transform}(SS_{out}^N) ),
    \end{aligned}
\end{equation}
where $N=\{ i\in \mathbb{R}: i=\{1,2,...,n\}\}$ represents the set of individual groups. MHSS maintains computational complexity comparable to standard selective scan~\cite{mamba} and provides advantages in parameter count and computational efficiency. Unlike 2DSS, which exhibits increased complexity with additional scanning directions, MHSS achieves improved computational efficiency via its multi-head strategy.

\subsubsection{All-Around Scanning}
\label{subsubsection::all-around}
Benefiting from the efficiency of MHSS, EAMamba introduces an all-around scanning strategy as the transformation function within MHSSM to enable holistic spatial dependency understanding. The strategy executes selective scanning across multiple directions: horizontal, vertical, diagonal, flipped diagonal, and their respective reversed orientations. Through this multi-directional scanning approach, all-around scanning strategy addresses the local pixel forgetting limitations inherent in the previous two-dimensional scanning strategy. Specifically, the all-around scanning pattern enables broader neighborhood information incorporation, which strengthens spatial context understanding and mitigates local pixel forgetting. As a result, this design enables MHSSM to achieve two objectives: computational efficiency through channel splitting and comprehensive visual information capture. Such a strategy facilitates effective image restoration without excessive computational overhead.

\section{Experimental Results}
\label{sec:experiment}
In this section, we present the experimental results.
We start with an overview of our experimental setup in Section~\ref{subsec:experimental_setup}. 
Following that, we evaluate our EAMamba on various image restoration tasks including denoising in Section~\ref{subsec:denoise}, super-resolution in Section~\ref{subsec:super_resolution}, deblurring in Section~\ref{subsec:deblur}, and dehazing in Section~\ref{subsec:dehaze}. 
Next, Section~\ref{subsec:effectiveness_scan} validates the effectiveness of various scanning strategies by visualizing their ERF results.
Finally, ablation studies for different design configurations of EAMamba are offered in Section~\ref{subsec:ablation_studies}.

\subsection{Experimental Setup}
\label{subsec:experimental_setup}

\paragraph{Architecture details.}
Our implementation adopts the following configuration as default parameters. EAMamba utilizes a four-level UNet architecture with [4, 6, 6, 7] MambaFormer blocks at respective levels. The refinement stage incorporates two MambaFormer blocks. The channel dimension $C$ maintains a constant value of 64. A simple feed-forward network~\cite{nafnet} is utilized as the default channel MLP.

\vspace{-10pt}
\paragraph{Training details.}
The training process includes 450,000 iterations with an initial learning rate of $3 \times 10^{-4}$, which decreases to $1 \times 10^{-6}$ through cosine annealing. The optimization employs AdamW~\cite{adamw} with parameters ($\beta_1=0.9, \beta_2=0.999$, weight decay $1e^{-4}$) and L1 loss. Following the progressive strategy in~\cite{restormer}, the training starts with patches of $128 \times 128$ pixels and batch size 64. These parameters progress to [(160, 40), (192, 32), (256, 16), (320, 8), (384, 8)] at iterations [138K, 234K, 306K, 360K, 414K], respectively. Our data augmentation includes random horizontal flipping, vertical flipping, as well as $90^\circ$ rotation.

\vspace{-10pt}
\paragraph{Measurement methods.}
The performance evaluation on benchmark datasets employs peak signal-to-noise ratio (PSNR) and structural similarity (SSIM) index measurements across RGB channels. For the RealSR dataset~\cite{realsr}, PSNR and SSIM computations use the Y channel in the YCbCr color space. We utilize the normalization approach to highlight subtle quality differences in visualization results between methods. The normalization against maximum difference values provides objective and standardized visual comparisons and addresses a fundamental challenge in fine-grained image difference presentation. The computational efficiency assessment adopts floating point operations per second (FLOPs) calculated through fvcore~\cite{fvcore} for all Vision Mamba methods, in accordance with the methodology in~\cite{vmamba} to ensure a fair comparison. The FLOPs are measured at a resolution of $256 \times 256$ for all experiments. Please note that in the tables presented, the best results among Vision Mamba-based approaches are highlighted in \bd{blue}, while the overall best results are highlighted in \st{red}.

\begin{figure*}[t]
    \centering
    \includegraphics[width=0.85\textwidth]{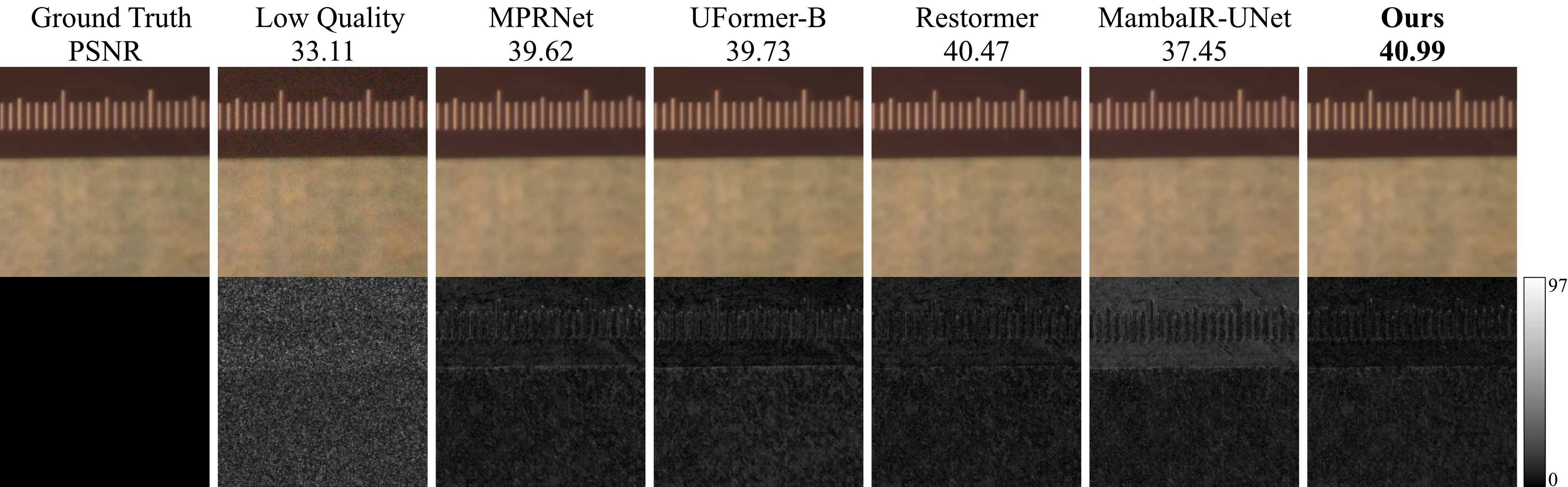}
    \caption{Visual comparison of MPRNet~\cite{mprnet}, UFormer-B~\cite{uformer}, Restormer~\cite{restormer}, MambaIR-UNet~\cite{mambair}, and our EAMamba on SIDD~\cite{sidd} validation set. The first row illustrates the results and the second row shows the normalized difference between the ground truth and the generated results.}
    \vspace{-3pt}
    \label{fig:sidd-compare}
\end{figure*}
\begin{figure*}[t]
    \centering
    \includegraphics[width=0.85\textwidth]{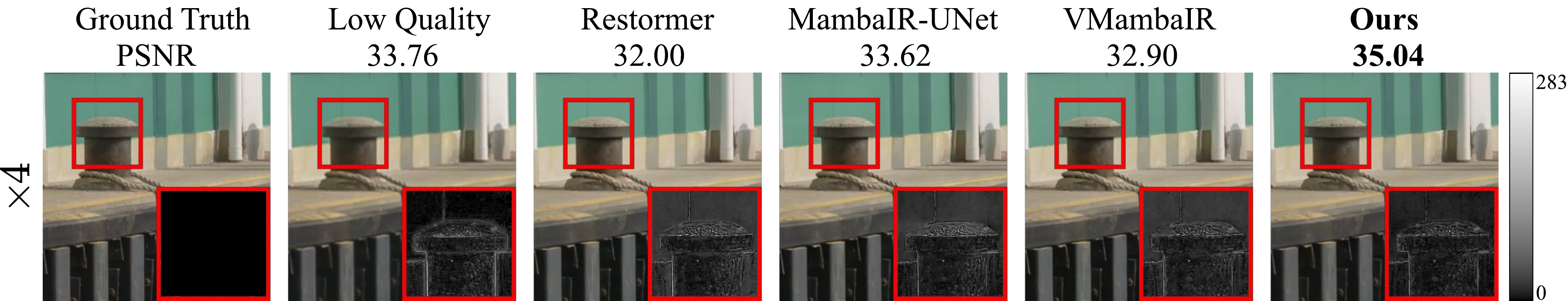}
    \caption{Visual comparison of Restormer~\cite{restormer}, MambaIR-UNet~\cite{mambair}, VmambaIR~\cite{vmambair}, and our EAMamba on the real-world super-resolution RealSR dataset~\cite{realsr} at scaling factors of $\times4$. The cropped regions represent the normalized difference between the ground truth and the generated results.}
    \vspace{-3pt}
    \label{fig:realsr-compare}
\end{figure*}

\subsection{Image Denoising}
\label{subsec:denoise}
\begin{table}[t]
\centering
\footnotesize
\vspace{5pt}
\caption{The average PSNR (dB) are measured on the synthetic Gaussian color image denoising benchmark datasets~\cite{bsd, mcmaster, kodak24}. The symbol $^*$ indicates methods that train separate models for each noise level, while methods without this symbol learn a single model to handle various noise levels. The results are obtained from the original manuscripts.}
\renewcommand{\arraystretch}{1.1}
\setlength{\tabcolsep}{2pt}
\resizebox{\linewidth}{!}{
\begin{tabular}{l|cc|ccc|ccc|ccc} 
\toprule[1pt]
\multirow{2}{*}{Method} & Param. & FLOPs & \multicolumn{3}{c|}{CBSD68~\cite{bsd}}    & \multicolumn{3}{c}{Kodak24~\cite{kodak24}}   & \multicolumn{3}{c}{McMaster~\cite{mcmaster}} \\
                        & (M) $\downarrow$ & (G) $\downarrow$ & $\sigma=15$ & $\sigma=25$ & $\sigma=50$ & $\sigma=15$ & $\sigma=25$ & $\sigma=50$ & $\sigma=15$ & $\sigma=25$ & $\sigma=50$  \\ 
\midrule
IRCNN~\cite{ircnn}          & -         & -         & 33.86         & 31.16         & 27.86         & 34.69         & 32.18         & 28.93         & 34.58         & 32.18         & 28.91        \\
FFDNet~\cite{ffdnet}        & -         & -         & 33.87         & 31.21         & 27.96         & 34.63         & 32.13         & 28.98         & 34.66         & 32.35         & 29.18        \\
DnCNN~\cite{dncnn}          & -         & -         & 33.90         & 31.24         & 27.95         & 34.60         & 32.14         & 28.95         & 33.45         & 31.52         & 28.62        \\
BRDNet$^*$~\cite{brdnet}    & -         & -         & 34.10         & 31.43         & 28.16         & 34.88         & 32.41         & 29.22         & 35.08         & 32.75         & 29.52        \\
DRUNet~\cite{drunet}        & 32.6      & 144       & 34.30         & 31.69         & 28.51         & 35.31         & 32.89         & 29.86         & 35.40         & 33.14         & 30.08        \\
SwinIR$^*$~\cite{swinir}    & \st{11.5} & 788       & 34.42         & 31.78         & 28.56         & 35.34         & 32.89         & 29.79         & 35.61         & 33.20         & 30.22        \\
Restormer~\cite{restormer}  & 26.1      & 141       & 34.39         & 31.78         & 28.59         & \st{35.44}    & \st{33.02}    & \st{30.00}    & 35.55         & 33.31         & 30.29        \\
\midrule
MambaIR$^*$~\cite{mambair}  & \bd{15.8} & 1290      & \st{34.43}    & 31.80         & 28.61         & 35.34         & 32.91         & 29.85         & \st{35.62}    & \bd{33.35}    & \st{30.31}    \\
EAMamba (Ours)              & 25.3      & \st{137}  & \st{34.43}    & \st{31.81}    & \st{28.62}    & \bd{35.36}    & \bd{32.95}    & \bd{29.91}    & 35.59         & 33.34         & \st{30.31}    \\
\bottomrule[1pt]
\end{tabular}
}
\label{table:denoise_results}
\vspace{-15pt}
\end{table}

In this evaluation, we evaluate EAMamba's performance on image denoising tasks, which encompass both synthetic Gaussian color denoising and real-world scenarios. Our comparison primarily focuses on the Vision Mamba baseline MambaIR~\cite{mambair}, following their evaluation protocol with two architectural variants: the vanilla model for synthetic Gaussian color denoising and MambaIR-UNet for real-world denoising tasks. In addition, we include results from several non-Mamba-based methods~\cite{ircnn, ffdnet, dncnn, brdnet, drunet, swinir, restormer, bm3d, cbdnet, ridnet, vdn, sadnet, danet, cycleisp, mirnet, deamnet, mprnet, dagl, ipt, uformer} for reference. For synthetic Gaussian color denoising, we train a single EAMamba model on noise levels $\sigma [0-50]$ using images from DIV2K~\cite{div2k}, Flickr2K~\cite{ntire}, WED~\cite{wed}, and BSD~\cite{bsd}. The evaluation employs benchmark datasets, CBSD68~\cite{bsd}, Kodak24~\cite{kodak24}, and McMaster~\cite{mcmaster}, across multiple noise levels $\sigma = [15, 25, 50]$. Table~\ref{table:denoise_results} presents quantitative results that demonstrate EAMamba's significant computational efficiency: it requires only $11\%$ of the FLOPs compared to MambaIR~\cite{mambair} with slight improvement in PSNR values. For real-world denoising, we train EAMamba and conduct evaluations on the SIDD dataset~\cite{sidd}. The quantitative results in Table~\ref{table:sidd_results} demonstrate that EAMamba reduces computational complexity by $41\%$ in FLOPs compared to MambaIR-Unet~\cite{mambair}, with only marginal PSNR decreases of 0.02 dB. Fig.~\ref{fig:sidd-compare} further presents qualitative results on the SIDD dataset, where EAMamba's generated images show closer correspondence to ground truth compared to the other methods. These experimental results conclusively validate EAMamba's efficiency and effectiveness in denoising tasks.

\begin{table}[t]
\centering
\footnotesize
\caption{The average PSNR (dB) and SSIM are measured on the real-world denoising benchmark SIDD~\cite{sidd}. The symbol $^*$ indicates methods using additional training data, while methods without the symbol only train with SIDD~\cite{sidd}. These baseline results are obtained from their original papers.}
\renewcommand{\arraystretch}{1.1}
\resizebox{0.95\linewidth}{0.17\textheight}{
\begin{tabular}{l|cc|cc} 
\toprule[1pt]
\multirow{2}{*}{Method} & Param. & FLOPs & \multicolumn{2}{c}{SIDD~\cite{sidd}} \\
                        & (M) $\downarrow$ & (G) $\downarrow$ & PSNR $\uparrow$ & SSIM $\uparrow$  \\ 
\midrule
DnCNN~\cite{dncnn}              & -      & -     & 23.66 & 0.583 \\
BM3D~\cite{bm3d}                & -      & -     & 25.65 & 0.685 \\
CBDNet$^*$~\cite{cbdnet}        & -      & -     & 30.78 & 0.801 \\
RIDNet$^*$~\cite{ridnet}        & \st{1.5} & 98  & 38.71 & 0.951 \\
VDN~\cite{vdn}                  & 7.8    & 44    & 39.28 & 0.956 \\
SADNet$^*$~\cite{sadnet}        & -      & -     & 39.46 & 0.956 \\
DANet$^*$~\cite{danet}          & 63.0   & \st{30} & 39.47 & 0.957 \\
CycleISP$^*$~\cite{cycleisp}    & 2.8    & 184   & 39.52 & 0.957 \\
MIRNet~\cite{mirnet}            & 31.8   & 785   & 39.72 & 0.959 \\
DeamNet$^*$~\cite{deamnet}      & 2.3    & 147   & 39.47 & 0.957 \\
MPRNet~\cite{mprnet}            & 15.7   & 588   & 39.71 & 0.958 \\
DAGL~\cite{dagl}                & 5.7    & 273   & 38.94 & 0.953 \\
HINet~\cite{hinet}              & 88.7   & 171   & 39.99 & 0.958 \\
IPT$^*$~\cite{ipt}              & 115.3  & 380   & 39.10 & 0.954 \\
MAXIM-3S~\cite{maxim}           & 22.2   & 339   & 39.96 & \st{0.960} \\
UFormer-B~\cite{uformer}        & 50.9   & 89    & 39.89 & \st{0.960} \\
Restormer~\cite{restormer}      & 26.1   & 141   & \st{40.02} & \st{0.960} \\
\midrule
MambaIR-UNet~\cite{mambair}     & 26.8      & 230      & \bd{39.89} & \st{0.960} \\
EAMamba (Ours)                  & \bd{25.3} & \bd{137} & 39.87      & \st{0.960} \\
\bottomrule[1pt]
\end{tabular}
}
\label{table:sidd_results}
\vspace{-15pt}
\end{table}

\subsection{Image Super-Resolution}
\label{subsec:super_resolution}
\begin{table}[t]
\centering
\footnotesize
\caption{The average PSNR (dB) and SSIM are measured on the RealSR dataset~\cite{realsr}. The results of ~\cite{restormer, mambair, vmambair} are obtained by training with the official repositories.}
\renewcommand{\arraystretch}{1.1}
\setlength{\tabcolsep}{2pt}
\resizebox{\linewidth}{!}{
\begin{tabular}{l|cc|cccccc}
\toprule[1pt]
\multirow{2}{*}{Method} & Param. & Flops & \multicolumn{2}{c}{x2} & \multicolumn{2}{c}{x3} & \multicolumn{2}{c}{x4}  \\
                        & (M) $\downarrow$ & (G) $\downarrow$ & PSNR $\uparrow$ & SSIM $\uparrow$ & PSNR $\uparrow$ & SSIM $\uparrow$ & PSNR $\uparrow$ & SSIM $\uparrow$ \\ 
\midrule
Restormer~\cite{restormer}  & 26.1  & 155   & \st{34.33}     & \st{0.929}     & \st{31.16}     & \st{0.874}     & 29.54     & \st{0.836}     \\
\midrule
MambaIR-UNet~\cite{mambair} & 26.8      & 230       & \bd{34.20}    & \bd{0.927}    & \st{31.16}    & \bd{0.872}    & 29.53         & 0.835         \\
VMambaIR~\cite{vmambair}    & 26.3      & 200       & 34.16         & \bd{0.927}    & 31.14         & \bd{0.872}    & 29.56         & \st{0.836}    \\
EAMamba (Ours)              & \st{25.3} & \st{137}  & 34.18         & \bd{0.927}    & 31.11         & \bd{0.872}    & \st{29.60}    & 0.835         \\
\bottomrule[1pt]
\end{tabular}
}
\label{table:realsr_results}
\vspace{-15pt}
\end{table}

In addition to denoising tasks, we compare EAMamba against MambaIR~\cite{mambair} and VMambaIR~\cite{vmambair} on the RealSR~\cite{realsr} benchmark dataset. Moreover, we present results from Restormer~\cite{restormer}, a non-Mamba-based architecture, as a reference. Table~\ref{table:realsr_results} summarizes parameter counts, computational FLOPs, and PSNR/SSIM results for scaling factors $\times2$, $\times3$, and $\times4$. Among all compared methods, EAMamba exhibits superior computational efficiency with the lowest parameter count and FLOPs.  EAMamba achieves superior PSNR performance at $\times4$ scaling and maintains a minimal PSNR gap (less than 0.05) at $\times2$ and $\times3$ scaling factors compared to other Vision Mamba methods. Fig.~\ref{fig:realsr-compare} further presents qualitative comparisons. The cropped difference results demonstrate EAMamba's capability in structural preservation and detail reconstruction, surpassing that of Vision Mamba baselines.
The above findings collectively substantiate EAMamba's advantages in super-resolution.

\begin{figure*}[t]
    \centering
    \includegraphics[width=0.85\textwidth]{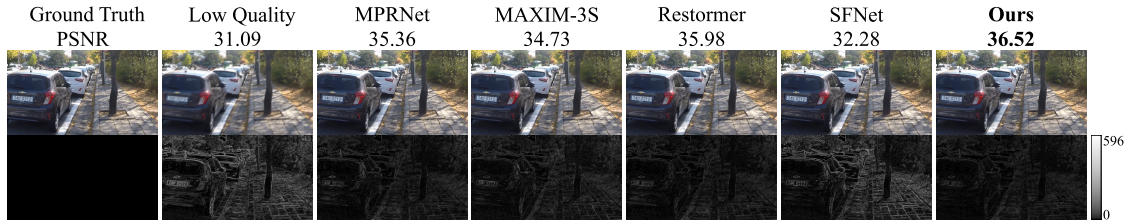}
    \caption{Visual comparison of MPRNet~\cite{mprnet}, MAXIM-3S~\cite{maxim}, Restormer~\cite{restormer}, SFNet~\cite{sfnet}, and our EAMamba on GoPro~\cite{gopro} validation set. The first row illustrates the results and the second row shows the normalized difference between the ground truth and the generated results.}
    \vspace{-3pt}
    \label{fig:gopro-compare}
\end{figure*}

\begin{figure*}[t]
    \centering
    \includegraphics[width=0.85\textwidth]{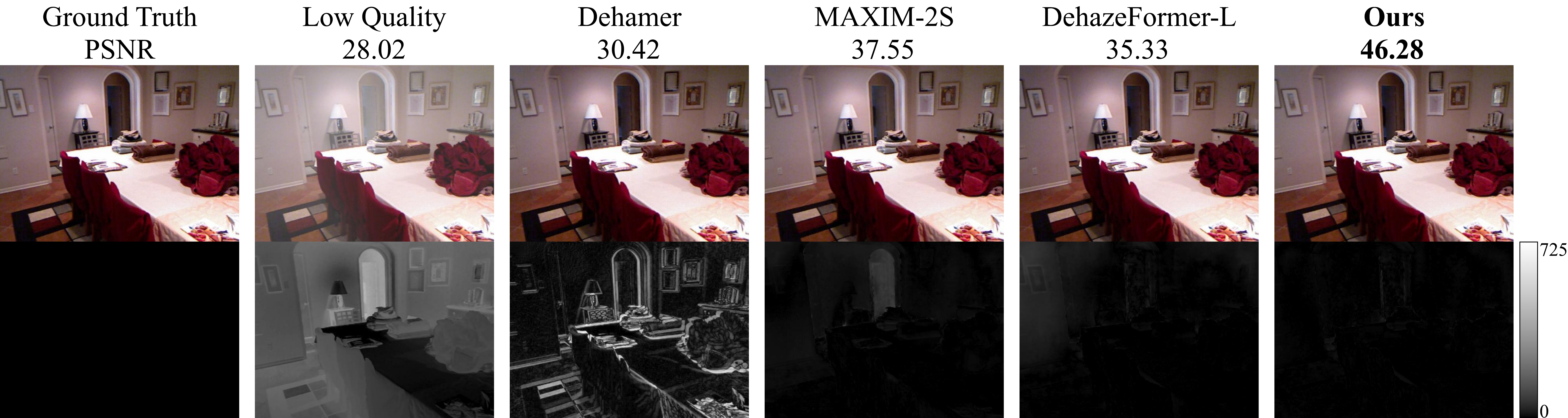}
    \caption{Visual comparison of Dehamer~\cite{dehamer}, MAXIM-2S~\cite{maxim}, DehazeFormer-L~\cite{dehazeformer}, and our EAMamba on SOTS-Indoor~\cite{reside} test set. The first row illustrates the results and the second row shows the normalized difference between the ground truth and the generated results}
    \vspace{-3pt}
    \label{fig:dehaze-compare}
\end{figure*}

\subsection{Image Deblurring}
\label{subsec:deblur}
\begin{table}[t]
\centering
\footnotesize
\caption{The average PSNR (dB) and SSIM are measured on the motion deblurring benchmark datasets~\cite{gopro, hide}. The results are obtained from the original manuscripts.}
\renewcommand{\arraystretch}{1.1}
\resizebox{\linewidth}{!}{
\begin{tabular}{l|cc|cc|cc}
\toprule[1pt]
\multirow{2}{*}{Method} & Param. & FLOPs & \multicolumn{2}{c|}{GoPro~\cite{gopro}} & \multicolumn{2}{c}{HIDE~\cite{hide}}  \\
                        & (M) $\downarrow$ & (G) $\downarrow$ & PSNR $\uparrow$ & SSIM $\uparrow$ & PSNR $\uparrow$ & SSIM $\uparrow$   \\ 
\midrule
DeblurGAN-v2~\cite{deblurganv2} & -     & -     & 29.55 & 0.934 & 26.61 & 0.875 \\
SRN~\cite{srn}                  & -     & -     & 30.26 & 0.934 & 28.36 & 0.915 \\
DBGAN~\cite{dbgan}              & -     & -     & 31.10 & 0.942 & 28.94 & 0.915 \\
DMPHN~\cite{dmphn}              & 21.7  & 195   & 31.20 & 0.940 & 29.09 & 0.924 \\
SPAIR~\cite{spair}              & -     & -     & 32.06 & 0.953 & 30.29 & 0.931 \\
MIMO-UNet+~\cite{mimo}          & 16.1  & 151   & 32.45 & 0.957 & 29.99 & 0.930 \\
MPRNet~\cite{mprnet}            & 20.1  & 760   & 32.66 & 0.959 & 30.96 & 0.939 \\
HINet~\cite{hinet}              & 88.7  & 171   & 32.71 & 0.959 & 30.32 & 0.932 \\
IPT~\cite{ipt}                  & 115.3 & 380   & 32.52 & -     & -     & -     \\
MAXIM-3S~\cite{maxim}           & 22.2  & 339   & 32.86 & 0.961 & \st{32.83} & \st{0.956} \\
UFormer-B~\cite{uformer}        & 50.9  & \st{89} & 33.06 & 0.967 & 30.90 & 0.953 \\
Restormer~\cite{restormer}      & 26.1  & 141   & 32.92 & 0.961 & 31.22 & 0.942 \\
Stripformer~\cite{stripformer}  & 19.7  & 155   & 33.08 & 0.962 & 31.03 & 0.940 \\
SFNet~\cite{sfnet}              & \st{13.3}  & 125   & 33.27 & 0.963 & 31.10 & 0.941 \\
\midrule
EAMamba (Ours)                  & \bd{25.3} & \bd{137}  & \st{33.58}    & \st{0.966}    & \bd{31.42}    & \bd{0.944}    \\
\bottomrule[1pt]
\end{tabular}
}
\label{table:deblur_results}
\vspace{-15pt}
\end{table}

To evaluate the performance of EAMamba in image deblurring tasks, we trained EAMamba using the GoPro dataset~\cite{gopro} and conducted evaluations on both the GoPro and HIDE benchmark datasets~\cite{gopro,hide}. The quantitative comparisons of our results with those from the other deblurring methods are presented in Table~\ref{table:deblur_results}. It is observed that EAMamba achieves superior performance on the GoPro benchmark, surpasses the second-best result by 0.31 dB in terms of PSNR, and ranks as the second-best on the HIDE benchmark. Fig.~\ref{fig:gopro-compare} offers qualitative comparisons across different deblurring methods. From the nearest object, such as the rear bumper, to the broader environmental context, including the brick pavement, EAMamba excels in preserving details. In the second row, EAMamba's predictions are nearly indistinguishable from the ground truth.
These results confirm EAMamba's capability in deblurring.

\subsection{Image Dehazing}
\label{subsec:dehaze}
\begin{table}[t]
\centering
\footnotesize
\caption{The average PSNR (dB) and SSIM are measured on the synthetic dehazing benchmark datasets~\cite{reside}. The results are obtained from the original manuscripts.}
\renewcommand{\arraystretch}{1.1}
\resizebox{\linewidth}{!}{
\begin{tabular}{l|cc|cc|cc}
\toprule[1pt]
\multirow{2}{*}{Method} & Param. & FLOPs & \multicolumn{2}{c|}{SOTS-Indoor~\cite{reside}} & \multicolumn{2}{c}{SOTS-Outdoor~\cite{reside}} \\
                        & (M) $\downarrow$ & (G) $\downarrow$ & PSNR $\uparrow$ & SSIM $\uparrow$ & PSNR $\uparrow$ & SSIM $\uparrow$ \\
\midrule
DehazeNet~\cite{dehazenet}          & -      & -     & 19.82 & 0.821    & 24.75 & 0.927 \\
AOD-Net~\cite{aodnet}               & -      & -     & 20.51 & 0.816    & 24.14 & 0.920 \\
GridDehazeNet~\cite{griddehazenet}  & \st{0.96} & \st{21} & 32.16 & 0.984    & 30.86 & 0.982 \\
MSBDN~\cite{msbdn}                  & 31.4   & 42    & 33.67 & 0.985    & 33.48 & 0.982 \\
FFA-Net~\cite{ffanet}               & 4.5    & 288   & 36.39 & 0.989    & 33.57 & 0.984 \\
ACER-Net~\cite{acernet}             & 2.6    & 52    & 37.17 & 0.990    & -     & -     \\
PMNet~\cite{pmnet}                  & 18.9   & 81    & 38.41 & 0.990    & 34.74 & 0.985 \\
MAXIM-2S~\cite{maxim}               & 14.1   & 216   & 38.11 & 0.991    & 34.19 & 0.985 \\
DeHamer~\cite{dehamer}              & 132.5  & 49    & 36.63 & 0.988    & 35.18 & 0.986 \\
DehazeFormer-L~\cite{dehazeformer}  & 25.4   & 280   & 40.05 & \st{0.996}    & -     & -     \\
\midrule
EAMamba (Ours)                      & \bd{25.3} & \bd{137}  & \st{43.19}    & \bd{0.995}    & \st{36.34}    & \st{0.988} \\
\bottomrule[1pt]
\end{tabular}
}
\label{table:dehaze_results}
\vspace{-10pt}
\end{table}

To further demonstrate EAMamba's restoration capabilities, we conduct dehazing experiments using the synthetic benchmark dataset RESIDE~\cite{reside} for both training and evaluation. The comparison includes ViT-based models~\cite{dehamer, dehazeformer} and other state-of-the-art architectures~\cite{dehazenet, aodnet, griddehazenet, msbdn, ffanet, acernet, maxim}. Table~\ref{table:dehaze_results} demonstrates EAMamba's balance between model complexity in terms of parameters and FLOPs, and image quality measured by PSNR and SSIM metrics. Fig.~\ref{fig:dehaze-compare} presents qualitative assessments. The evaluation results reveal that EAMamba exhibits superior detail preservation with minimal deviation from ground truth. These findings validate EAMamba's efficiency and effectiveness in dehazing tasks.

\begin{figure}[t]
    \centering
    \includegraphics[width=\columnwidth]{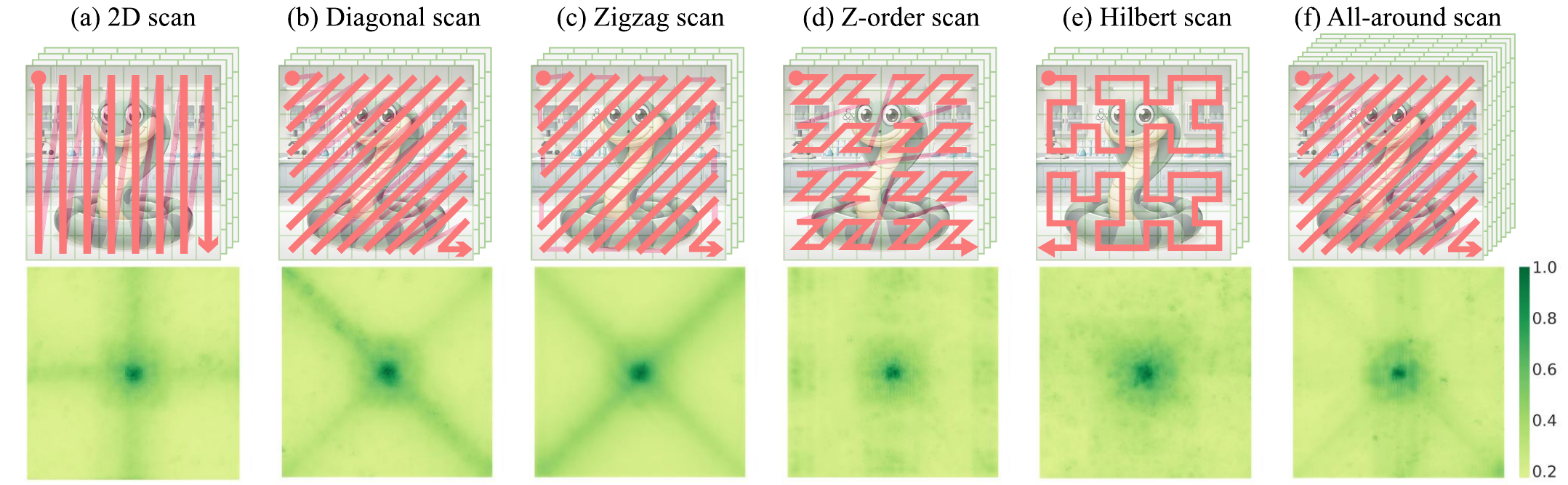}
    \caption{Illustration of the ERF results for different scanning strategies, including two-dimensional scan, diagonal scan, zigzag scan, Z-order scan, Hilbert scan, and our all-around scan with reversing and flipping.}
    \label{fig:erf}
    \vspace{-10pt}
\end{figure}

\subsection{Effectiveness of Various Scanning Strategies}
\label{subsec:effectiveness_scan}
To validate the effectiveness of our all-around scanning strategy, we visualize the ERF results of different scanning strategies in Fig.~\ref{fig:erf}. The ERF of the 2D scan, as shown in Fig.~\ref{fig:erf}~(a), demonstrates its capability to capture long-range dependencies in both horizontal and vertical directions. The diagonal and zigzag scans, shown in Figs.\ref{fig:erf}(b) and(c) respectively, excel at capturing global information along diagonal paths. The Hilbert scan's ERF, as shown in Fig.~\ref{fig:erf}~(e), lacks global information capture, while the Z-order scan, as displayed in Fig.~\ref{fig:erf}~(d), gathers global information with discontinuities. These approaches capture only specific spatial information. 
To capture complementary spatial information often missed by a single scanning strategy, we introduce an all-around scanning mechanism that combines multiple scanning strategies. For instance, we combine the 2D scan with the diagonal scan, as depicted in Fig.~\ref{fig:erf}~(f), which captures both the global and local contextual information from images. Moreover, the quantitative results in Table~\ref{table:scan} reveal that the all-around scanning strategy improves PSNR by $0.07$-$0.15$ dB over the other methodologies on the SIDD dataset~\cite{sidd}. As shown in Table~\ref{table:multiscan}, the combination of the 2D scan and the diagonal scan generally yields good performance and is set as the default configuration. Other combinations can be employed for specific use cases. The all-around scanning mechanism not only solves image restoration tasks by integrating multiple scanning strategies, but also offers the flexibility to seamlessly incorporate novel scanning strategies with MHSSM.

\begin{table}[t]
\centering
\footnotesize
\caption{The average PSNR (dB) on SIDD~\cite{sidd} for various scan strategies.}
\renewcommand{\arraystretch}{1.1}
\setlength{\tabcolsep}{15pt}
\resizebox{\linewidth}{!}{
\begin{tabular}{c|c|c|c|c|c}
\toprule[1pt]
2D    & Diagonal & Zigzag & Z-order & Hilbert & All-around  \\
\midrule
39.80 & 39.79    & 39.77  & 39.74   & 39.74   & \st{39.87}  \\
\bottomrule[1pt]
\end{tabular}
}
\vspace{-2pt}
\label{table:scan}
\end{table}
\begin{table}[t]
\centering
\footnotesize
\caption{The average PSNR (dB) on various image restoration datasets~\cite{sidd, realsr, gopro, reside}  with different combination of scanning strategies.}
\resizebox{\linewidth}{!}{
\begin{tabular}{l|cccc}
\toprule[1pt]
Dataset     & 2D + Diagonal & 2D + Z-order & 2D + Hilbert & 2D + Diagonal + Z-order   \\ 
\midrule
SIDD~\cite{sidd}        & \st{39.87}    & 39.82         & 39.83         & 39.83         \\
RealSRx4~\cite{realsr}      & \st{29.60}    & 29.58         & 29.51         & 29.57         \\
GoPro~\cite{gopro}          & 33.58         & 33.51         & \st{33.66}    & 33.56         \\
SOTS-Indoor~\cite{reside}   & 43.19         & 43.20         & 43.07         & \st{43.37}    \\
\bottomrule[1pt]
\end{tabular}
}
\vspace{-2pt}
\label{table:multiscan}
\end{table}

\vspace{-5pt}
\subsection{Ablation Studies}
\label{subsec:ablation_studies}
\vspace{-5pt}
\paragraph{The effectiveness of MHSS and all-around scan.}
\begin{table}[t]
\centering
\footnotesize
\caption{The average PSNR (dB) on Urban100~\cite{urban100} with Gaussian color image denoising for different design choices.}
\renewcommand{\arraystretch}{1.1}
\resizebox{0.85\linewidth}{!}{
\begin{tabular}{l|cc|ccc}
\toprule[1pt]
\multirow{2}{*}{Method}   & Param. & FLOPs & \multicolumn{3}{c}{Urban100~\cite{urban100}}            \\
                          & (M) $\downarrow$ & (G) $\downarrow$ & $\sigma=15$ & $\sigma=25$ & $\sigma=50$  \\
\midrule
Baseline            & 31.1      & 286       & \st{35.15}    & \st{33.00}    & \st{30.08}    \\
+ MHSSM             & \st{25.3} & \st{137}  & 35.06         & 32.89         & 29.95         \\
+ all-around scan   & \st{25.3} & \st{137}  & 35.10         & 32.93         & 30.01         \\
\bottomrule[1pt]
\end{tabular}
}
\label{table:component}
\vspace{-15pt}
\end{table}
To validate EAMamba's architectural decisions, Table~\ref{table:component} quantifies the contribution of each design element. The baseline model utilizes 2DSSM~\cite{mambair} with 2D scanning. Our analysis reveals that substituting 2DSSM by MHSSM reduces the computational cost in FLOPs by half, while only resulting in a negligible $0.1\%$ decrease in PSNR. This outcome quantitatively demonstrates MHSSM's capacity for significant computational savings with minimal detriment to the quality of image restoration. Moreover, Table~\ref{table:component} suggests that the all-around scanning delivers enhanced image quality compared to conventional 2D scanning. These results support the design choices in EAMamba and highlights its ability to balance efficiency with restoration performance.

\vspace{-15pt}
\paragraph{Comparison of the various channel MLP.}
\begin{table}[t]
\centering
\footnotesize
\caption{The average PSNR (dB) on Urban100~\cite{urban100} with Gaussian color image denoising for different channel MLPs choices.}
\renewcommand{\arraystretch}{1.1}
\resizebox{0.85\linewidth}{!}{
\begin{tabular}{l|cc|ccc}
\toprule[1pt]
\multirow{2}{*}{Channel MLP} & Param. & FLOPs & \multicolumn{3}{c}{Urban100~\cite{urban100}}             \\
                             & (M) $\downarrow$ & (G) $\downarrow$ & $\sigma=15$ & $\sigma=25$ & $\sigma=50$  \\ 
\midrule
None                         & 16.5         & 90        & 34.98         & 32.79         & 29.82         \\
\midrule
FFN                          & 28.3         & 153       & 35.10         & 32.93         & 30.01         \\
GDFN                         & 34.5         & 189       & \st{35.15}    & \st{32.98}    & \st{30.08}    \\
Simple FFN                   & \st{25.3}    & 137       & 35.10         & 32.93         & 30.01         \\
CA                           & 28.3         & \st{123}  & 35.05         & 32.88         & 29.95         \\
\bottomrule[1pt]
\end{tabular}
}
\label{table:ffn}
\vspace{-18pt}
\end{table}

Table~\ref{table:ffn} evaluates various channel MLP modules within the MambaFormer Block, as discussed in Section~\ref{subsec::mambaformer}, including vanilla Feed-Forward Network (FFN)~\cite{transformer}, Gated-Dconv FFN (GDFN)~\cite{restormer}, Simple FFN~\cite{nafnet}, and Channel Attention (CA)~\cite{rcan}. The exclusion of Channel MLP results in a performance decline greater than $0.1$ dB in PSNR. While GDFN demonstrates superior performance, Simple FFN achieves an optimal balance between performance and computational efficiency. This balance motivates our selection of Simple FFN as the default configuration. More detailed specifications are provided in the supplementary materials.

\vspace{-20pt}
\section{Conclusion}
\vspace{-5pt}
\label{sec:conclusion}
This paper presented EAMamba, an enhanced Vision Mamba framework that advances the architectural design of previous Vision Mamba approaches. EAMamba introduces two architectural innovations: MHSSM and the all-around scanning mechanism. MHSSM enhances framework scalability through channel-split feature processing, which enables efficient aggregation of flattened one-dimensional sequences and improved computational efficiency in Vision Mamba frameworks. The all-around scanning mechanism extends beyond conventional two-dimensional scanning strategies to capture comprehensive spatial information, which effectively resolves the local pixel forgetting limitation. Our extensive experimental evaluations across multiple image restoration tasks, including image super resolution, denoising, deblurring, and dehazing, validated these architectural innovations. Both qualitative and quantitative results demonstrate that EAMamba achieves favorable performance with reduced parameters and FLOPs compared to existing low-level Vision Mamba methods.
\vspace{-5pt}
\section*{Acknowledgements}
\vspace{-5pt}

The authors gratefully acknowledge the support of the National Science and Technology Council in Taiwan under grant numbers MOST 111-2223-E-002-011-MY3, NSTC 113-2221-E-002-212-MY3, and NSTC 113-2640-E-002-003. The authors also appreciate the donation of GPUs from NVIDIA Corporation and the support from MediaTek Inc. Additionally, the authors extend their gratitude to the National Center for High-Performance Computing for providing computational resources.

\small \bibliography{main}

\end{document}